\documentclass[a4paper]{article}
\usepackage{INTERSPEECH2022}

\usepackage{amsmath,graphicx}

\usepackage{caption}
\usepackage{wrapfig}
\usepackage{setspace}
\usepackage{cleveref}
\usepackage{enumitem}

\usepackage{multirow}
\usepackage{multicol}
\usepackage{xcolor}
\usepackage{esvect}

\makeatletter
\renewcommand{\section}{\@startsection
  {section}%
  {1}%
  {}%
  {-0.5\baselineskip}%
  {0.2\baselineskip}%
  {}}%

\renewcommand{\subsection}{\@startsection
  {subsection}%
  {2}%
  {}%
  {-0.1\baselineskip}%
  {0.1\baselineskip}%
  {}}%

\renewcommand{\subsubsection}{\@startsection
  {subsubsection}%
  {3}%
  {}%
  {-0.2\baselineskip}%
  {0.2\baselineskip}%
  {}}%

\g@addto@macro\normalsize{%
  \setlength\abovedisplayskip{5pt plus 2pt minus 2pt}
  \setlength\belowdisplayskip{5pt plus 2pt minus 2pt}
  \setlength\abovedisplayshortskip{4pt plus 2pt minus 2pt}
  \setlength\belowdisplayshortskip{4pt plus 2pt minus 2pt}
}

\makeatother

\Crefname{equation}{Eq.}{Eqs.}
\Crefname{figure}{Fig.}{Figs.}
\Crefname{tabular}{Tab.}{Tabs.}

\DeclareMathOperator*{\argmax}{arg\,max\hspace{2mm}}

\newcommand\numberthis{\addtocounter{equation}{1}\tag{\theequation}}

\def\L{{\cal L}}
\def\B{{\cal B}}
\def\W{{\cal W}}
\def\H{{\cal H}}

\title{Efficient Training of Neural Transducer for Speech Recognition}
\name{Wei Zhou, Wilfried Michel, Ralf Schl\"uter, Hermann Ney}
\address{
  Human Language Technology and Pattern Recognition, Computer Science Department,\\
  RWTH Aachen University, 52074 Aachen, Germany \\
  AppTek GmbH, 52062 Aachen, Germany}
\email{\{zhou,michel,schlueter,ney\}@cs.rwth-aachen.de}

\begin{document}

\maketitle
\begin{abstract}
As one of the most popular sequence-to-sequence modeling approaches for speech recognition, the RNN-Transducer has achieved evolving performance with more and more sophisticated neural network models of growing size and increasing training epochs.
While strong computation resources seem to be the prerequisite of training superior models, we try to overcome it by carefully designing a more efficient training pipeline.
In this work, we propose an efficient 3-stage progressive training pipeline to build highly-performing neural transducer models from scratch with very limited computation resources in a reasonable short time period.
The effectiveness of each stage is experimentally verified on both Librispeech and Switchboard corpora. 
The proposed pipeline is able to train transducer models approaching state-of-the-art performance with a single GPU in just 2-3 weeks. 
Our best conformer transducer achieves 4.1\% WER on Librispeech test-other with only 35 epochs of training.
\end{abstract}

\vspace{-1.5mm}
\section{Introduction}
Recently, sequence-to-sequence modeling becomes the major trend of automatic speech recognition (ASR).
Among other popular approaches such as connectionist temporal classification (CTC) \cite{graves2016ctc} and attention-based encoder-decoder (AED) models \cite{bahdanau2016end, chan2016listen}, the recurrent neural network transducer (RNN-T) \cite{graves2012rnnt} receives the most interest due to its good performance and streaming nature.

The RNN-T model definition allows a direct from-scratch training by summing over all alignment paths matching the output sequence.
This simplicity usually comes at the cost of high memory and computation requirement.
Meanwhile, the performance of transducer models has largely evolved with more and more sophisticated neural network (NN) architectures \cite{zhang20trafoTransducer, han2020contextNet, Gulati20conformer}.
The best performing systems usually adopt a large NN trained for many epochs.
All of these seem to indicate that strong computation resources have become the prerequisite of training highly-performing transducer models in a reasonable time.
Although this can be true to some extent, we try to overcome it by carefully designing a more efficient training pipeline.

More specifically, we propose a fast 3-stage progressive training pipeline for neural transducer models in this work. We provide detailed recipes and design principles of each stage, which largely reduce the time and computation costs.
Experiments on both Librispeech (LBS) \cite{libsp} and Switchboard (SWB) \cite{swb} corpora show that our proposed pipeline is able to train transducer models approaching state-of-the-art (SOTA) performance from scratch with a single GPU in just 2-3 weeks.

\vspace{-1mm}
\section{Model}
In this work, we mainly focus on the strictly monotonic version of RNN-T \cite{tripathi2019monoRNNT} which closely matches the nature of speech as desired for many real-time applications, although most of the proposed approaches can also work for standard RNN-T. 
Let $X$ denote the acoustic feature sequence of a speech utterance and $h_1^T = f^{\text{enc}}(X)$ denote the encoder output, which transforms the input into high-level representations. Let $a_1^S$ denote the output label sequence of length $S \le T$, 
whose sequence posterior is defined as:\\
\scalebox{0.9}{\parbox{1.11\linewidth}{%
\begin{align*}
P_{\text{RNNT}}(a_1^S|X) &= \sum_{y_1^T: \B^{-1}(a_1^S)} P_{\text{RNNT}}(y_1^T | h_1^T) \\
&= \sum_{y_1^T:\B^{-1}(a_1^S)} \prod_{t=1}^{T} P_{\text{RNNT}}(y_t | \B(y_1^{t-1}), h_1^T) \numberthis \label{eq:rnnt}
\end{align*}}}
Here $y_1^T$ is the blank $\epsilon$-augmented alignment sequence, which is uniquely mapped to $a_1^S$ via the collapsing function $\B$ to remove all blanks. A limited context dependency \cite{zhou2021phonemeTransducer, ghodsi20statelessRNNT, prabhavalkar21rnntLessCtx} can also be  introduced to further simplify the model in \Cref{eq:rnnt}. This can be better expressed from a lattice representation of the transducer label topology \cite{graves2012rnnt, tripathi2019monoRNNT}. 
By denoting $y_1^{t-1}$ as a path reaching a node $(t-1, s-1)$, we have:\\
\scalebox{0.9}{\parbox{1.11\linewidth}{%
\begin{align*}
P_{\text{RNNT}}(y_t | \B(y_1^{t-1}), h_1^T) = P_{\text{RNNT}}(y_t | a_1^{s-1}, h_t)
= P_{\text{RNNT}}(y_t | a_{s-k}^{s-1}, h_t) 
\end{align*}}}
where $k$ is the label context size, and $y_1^t$ reaches $(t, s-1)$ if $y_t=\epsilon$ or $(t, s)$ otherwise.

For decoding using the RNN-T model with an external language model (LM), a general formulation of the maximum a posteriori (MAP) decision rule can be given as:\\
\scalebox{0.9}{\parbox{1.11\linewidth}{%
\begin{align*}
X \rightarrow \widehat{\W(a_1^S)} = \argmax_{\W(a_1^S, S)} P^{{\lambda}_1}_{\text{LM}}(\W(a_1^S)) \cdot \frac{P_{\text{RNNT}}(a_1^S|X)}{P^{{\lambda}_2}_{\text{RNNT-ILM}}(a_1^S)} \numberthis \label{eq:MAP-ILM}
\end{align*}}}
where $P_{\text{RNNT-ILM}}$ is the internal LM (ILM) / sequence prior of $P_{\text{RNNT}}$. 
Here ${\lambda}_1$ and ${\lambda}_2$ are scales applied in common practice, where ${\lambda}_2=0$ leads to the simple shallow fusion (SF) approach \cite{gulcehre2015shallowFusion}.
The additional function $\W$ is introduced for a more general coverage of different label choices for $a$. 
More specifically, $\W$ is the lexical mapping or identity function when $a$ represents phonemes or (sub)words, correspondingly.


\vspace{-1mm}
\section{Training Pipeline \& Recipes}
In a standard pipeline, the encoder of transducer NN is usually pretrained with the CTC \cite{graves2016ctc} objective, which can be time consuming when the encoder size grows large. The complete transducer NN is then trained with the full-sum (FS) loss:\\
\scalebox{0.9}{\parbox{1.11\linewidth}{%
\begin{align*}
\L_{\text{FS}} = -\log P_{\text{RNNT}}(a_1^S|X)
\numberthis \label{eq:fullsum}
\end{align*}}}
which is usually very time consuming w.r.t. computation cost and convergence to optimal performance. It also has a high memory consumption due to the FS over all alignment paths.

To further improve the FS-trained transducer model, the minimum Bayes risk (MBR) sequence discriminative training is commonly applied due to its consistency with the ASR evaluation metric.
For simplicity, we denote $a_1^S$ as $\vec{a}$ and omit the variable length $S$ for the MBR criterion:\\
\scalebox{0.9}{\parbox{1.11\linewidth}{%
\begin{align*}
\L_{\text{MBR}} = \sum_{\vec{a} \in \H} \frac{P_{\text{seq}}(\vec{a} | X)}{\sum_{\vec{a'} \in \H} P_{\text{seq}}(\vec{a'} | X)} \cdot R(\vec{a}, \vec{a}^r) \numberthis \label{eq:mbr}
\end{align*}}}\vspace{-0.5mm}
Here $\H$ is the hypotheses space, which is usually approximated with an $N$-best list.
The risk function $R$ is typically the Levenshtein distance w.r.t. the ground truth sequence $\vec{a}^r$. 
The $P_{\text{seq}}$ is often chosen to be $P_{\text{RNNT}}$ directly \cite{guo2020EfficientMBRtransducer, weng2020MBRtransducer}, but may also include different LMs as in \Cref{eq:MAP-ILM} \cite{zhong2021ilmMBR}. 
This objective requires generating $\H$ by (semi-) on-the-fly decoding with the model in training, and computing FS for all label sequences in $\H$.
Therefore, the MBR training has even much higher complexity and cost in terms of both time and memory.

In this section, we present the proposed 3-stage progressive training pipeline with detailed recipes. All hyper-parameters are extensively tuned on LBS only, which generalize well on other public corpora such as SWB and TED-LIUM Release 2 \cite{tedlium2}.
This efficient pipeline allows to train highly-performing transducer models from scratch with very limited computation resources in a reasonable short time period. 
We use conformer \cite{Gulati20conformer} transducer throughout this work, while the 
recipes also generalize well on common NN structures
such as bidirectional long short-term memory \cite{hochreiter1997lstm} (BLSTM).

\subsection{Stage 1: Viterbi training}
As a first step, we directly apply the frame-wise cross-entropy (CE) training to a from-scratch initialized transducer NN:\\
\scalebox{0.9}{\parbox{1.11\linewidth}{%
\vspace{-1mm}
\begin{align*}
\L_{\text{Viterbi}} = -\log P_{\text{RNNT}}(\hat{y}_1^T | h_1^T) = \sum_{t=1}^T -\log P_{\text{RNNT}}(\hat{y}_t | \B(\hat{y}_1^{t-1}), h_1^T)
\end{align*}}}
where the Viterbi alignment $\hat{y}_1^T$ is obtained by training a very small CTC model.
For label loops in the CTC alignment, we simply take the last frame of the segment as the label position.

As shown in \cite{zhou2021phonemeTransducer, zeyer2020transducer}, such Viterbi training for transducer allows an easy integration of various regularization methods and auxiliary losses to improve performance and stability:
\vspace{-1mm}
\begin{itemize}[leftmargin=*, itemsep=-0.3mm]
\item 0.2 label smoothing \cite{szegedy2016labelsmooth} on $\L_{\text{Viterbi}}$
\item $\L_{\text{enc}} = \sum_{t=1}^T -\log P(\hat{y}_t | h_t)$: an auxiliary CE loss w.r.t. $\hat{y}_1^T$ to the encoder output by introducing an additional softmax layer only in training, which is further weighted by a 1.0 focal loss factor \cite{lin2017focalloss}. The same can also be applied to the encoder middle layer with a smaller scale (default 0.3) \cite{zeineldeen2022hybridconformer}.
\item $\L_{\text{boost}}$: another auxiliary loss to boost the importance of output label sequence by simply excluding all blank frames in $\L_{\text{Viterbi}}$:\\
\scalebox{0.9}{\parbox{1.11\linewidth}{%
\vspace{-1mm}
\begin{align*}
\L_{\text{boost}} = \sum_{t=1}^T \hspace{1mm} -\log P_{\text{RNNT}}(\hat{y}_t | \B(\hat{y}_1^{t-1}), h_1^T) \cdot (1- \delta_{\hat{y}_t, \epsilon})
\end{align*}}}\vspace{-0.5mm}
By default, we further apply a scale $\alpha = 5$ to this loss.
\item chunking: each utterance can also be split into windows with 50\% overlap, where the window length can be adjusted based on the label context size of the model. 
\end{itemize}
\vspace{-1.2mm}
The above is our default recipe which yields the final loss as:\\
\scalebox{0.9}{\parbox{1.11\linewidth}{%
\vspace{-1mm}
\begin{align*}
\L^{\text{final}}_{\text{Viterbi}} = \L_{\text{Viterbi}} + \L_{\text{enc}} + \alpha\L_{\text{boost}}
\end{align*}}}
We also add a gradient clipping of 20 to ensure stability.
We observe that this Viterbi training is very robust to varying quality of the alignment CTC model. 
The simplified criterion allows an easy learning, which gives fast convergence to good performance with only a small number of epochs.

In contrast to the standard pipeline, we obtain several gains on speed and memory in this stage:\\
1. We only need to train a very small CTC model till some reasonable performance to generate the Viterbi alignment.\\
2. The Viterbi training itself is much faster than FS (3 times faster in our setup) and needs less number of epochs to reach a similar good performance.\\
3. Much less memory is needed, which allows a much larger batch size. Besides speed-up, this also leads to a more reliable BatchNorm operation in the conformer model.

However, after quickly reaching a good local optimum, this Viterbi training barely improves further even with much longer training.
This is mainly due to the limitation of fixed path training, which restricts the model to further converge to a better optimum. 
The problem can only be slightly eased with realignment, which nevertheless makes the training less efficient.


\subsection{Stage 2: full-sum fine-tuning}
As a result, we directly switch to FS fine-tuning using \Cref{eq:fullsum} as the 2nd stage after the model quickly converges in the Viterbi training stage. 
Without the aforementioned restriction, it gives the model more freedom to converge to a better optimum, which is also more consistent with the FS nature of transducer model definition.
Since the starting point is already a good base model, this stage only needs a very small amount of epochs to achieve further decent improvement.
This largely reduces the time cost of the FS training in the standard pipeline.
Due to memory consumption, we have to accumulate gradients from several smaller batches to form a usual-size mini-batch for update. Therefore, we freeze the parameters in BatchNorm operation from this stage on, which is also more consistent with testing.

\subsection{Stage 3: fast sequence training for LM integration}
\label{sec:fastMBR}
\vspace{-0.3mm}
\subsubsection{Objective}
\vspace{-0.5mm}
We also apply MBR sequence discriminative training as the 3rd stage to further improve the transducer model from stage 2. 
We aim to directly prepare the transducer model for external LM integration in this stage. 
As explained in \cite{zhou2022transducerILM}, FS-trained transducer models usually have a quite high blank probability, which forces a rather small scale ${\lambda}_1$ for the external LM in SF decoding.
Otherwise, the results will be biased towards shorter sequences with high deletion (Del) errors.
This weakens the LM's discrimination power, which is usually handled in decoding by applying ILM correction \cite{variani2020hat, ILME, zhou2022transducerILM} and/or heuristics such as length reward \cite{McDermott2019DR, saon21rnnt}.
However, such approaches require very careful tuning and a high quality of ILM estimation, which can be more problematic for transducers with limited context.

Therefore, we tend to directly tackle the fundamental issue by fine-tuning the transducer model with the MBR criterion towards an LM-aware distribution that further suppresses blank / boosts label around label positions while maintains the blank dominance elsewhere. 
Ideally, even with SF decoding, this allows a reasonably larger ${\lambda}_1$ to suppress substitution (Sub) and/or insertion (Ins) errors, while Del errors stay the same or decrease further. 
This should also make the scale-tuning easier for ILM correction if it is still necessary.
Thus, we apply LM SF 1-pass decoding to generate the $N$-best list. We optimize ${\lambda}_1$ on the dev set using the base transducer model from stage 2, which usually gives higher Del errors than Ins errors. The $P_{\text{seq}}$ in the MBR criterion of \Cref{eq:mbr} is then defined accordingly:\\
\scalebox{0.9}{\parbox{1.11\linewidth}{%
\vspace{-1mm}
\begin{align*}
P_{\text{seq}}(\vec{a} | X) = \left( P_{\text{RNNT}}(a_1^S|X) \cdot  P^{{\lambda}_1}_{\text{LM}}(\W(a_1^S)) \right)^{\beta}
\end{align*}}}\vspace{-1mm}
where $\beta$ is the global renormalization scale (typically $\beta=\frac{1}{{\lambda}_1}$ in our setup) and $P_{\text{LM}}$ is frozen in the training.
We always include $\vec{a}^r$ into the $N$-best and adopt 0.05-0.1 FS loss for stability, which yields the final loss as:\\
\scalebox{0.9}{\parbox{1.11\linewidth}{%
\vspace{-1mm}
\begin{align*}
\L^{\text{final}}_{\text{MBR}} = \L_{\text{MBR}} + \alpha \L_{\text{FS}}
\end{align*}}}

\vspace{-1.5mm}
\subsubsection{Static $N$-best generation}
\vspace{-0.5mm}
In \cite{guo2020EfficientMBRtransducer}, it is observed that a small $N$-best list usually does not change much within a certain training period, which allows a semi-on-the-fly $N$-best decoding using every new sub-epoch model.
Similarly, we also generate $N$-best separately and recompute the exact FS of $P_{\text{RNNT}}$ for each sequence in the $N$-best in training.
We observe that even with such semi-on-the-fly decoding, most improvement of the MBR training only comes in the first quarter of one full epoch.
Based on these, we further simplify the pipeline by only using the stage-2 base model to decode random 25\% of data for MBR training in this stage.

More specifically, we generate lattices for this subset of data to obtain the $N$-best list.
For reuse purpose, the lattices can be slightly larger to allow an $N$-best size increment and to contain more unique label sequences after duplicates removal, e.g. homophones when $a$ represents phonemes.
This becomes a once-only and completely offline process which can be parallelized to many CPUs or a few GPUs if available.
Different tuning experiments can then reuse the same lattices and finish within a few hours.
Additionally, this lattice generation process can be further split into several sub-epochs, where the first training experiment can start as soon as the first sub-epoch of data is decoded.
In this case, training and decoding can run in parallel with minor synchronization effort.
This whole process has a strong similarity to the lattice-based MBR training in the classical hybrid systems \cite{vesely2013hybridSeq, zhou2020tlv2}. 

Our $N$-best-based MBR training allows a dynamic $N$ to account for lattices containing less hypotheses than $N$, which can happen for some rather simple or short utterances.
We also find that it is safe to use a slightly weaker LM with reduced context/model size to further speed up the lattice generation process.
However, this weaker LM can not be too distinct from the final recognition LM so that the small $N$-best list still contains representative error patterns.
With a largely reduced complexity, the 3rd stage MBR training gives some further improvement for LM integration within a very short time, which completes the proposed training pipeline.


\subsection{Learning rate scheduling \& epochs}
\vspace{-0.5mm}
Inspired by \cite{saon21rnnt}, we also apply the one-cycle learning rate (OCLR) policy \cite{smith2019OCLR} to enhance the effectiveness of our training pipeline. We try to more closely study the original OCLR design and adapt it to a simple formulation that requires least tuning effort and works very well in general with common adaptive optimizer. 
This can be expressed as a 3-phase LR scheduling within one training stage:
\vspace{-0.8mm}
\begin{enumerate}[leftmargin=6mm , itemsep=-0.4mm]
\item [P1.] 0-45\% of total steps: linear increase LR from $\text{LR}_1$ to $\text{LR}_\text{peak}$ with $\text{LR}_1 = \text{LR}_\text{peak} / 10$
\item [P2.] 45-90\% of total steps: linear decrease LR from $\text{LR}_\text{peak}$ to $\text{LR}_2$ with $\text{LR}_2 = \text{LR}_\text{peak} / 10$
\item [P3.] 90-100\% of total steps: linear decrease LR from $\text{LR}_2$ to $\text{LR}_\text{final}$ with $\text{LR}_\text{final} = 1e^{-6}$ typically
\end{enumerate}
\vspace{-0.8mm}
As claimed in \cite{smith2019OCLR}, P1 and P2 share the same spirit as curriculum learning \cite{bengio2009CurriculumLearning} and simulated annealing \cite{arats1990annealing}, respectively. And P3 tries to find the optima of local optimum with a quick decay to a much smaller LR.
With this schedule, only $\text{LR}_\text{peak}$ needs to be tuned, whose optimum is usually very close to the peak LR used in a common constant + exponential decay policy.
We observe that the described OCLR schedule gives consistently better performance than the common one under the same training epochs. 
For stage 1 training, we use this OCLR schedule exactly.
For stage 2, we adjust P1 with $\text{LR}_1 = \text{LR}_\text{peak}$ and P2 with $\text{LR}_2 = \text{LR}_\text{peak} / 5$.
For stage 3, we simply use a constant LR $1e^{-5}$.
By default, we use a mini-batch size of 10-15k input frames, $5e^{-6}$ L2 regularization and 0.1 dropout, which works generally well with this schedule.
In terms of epochs, we find that a reasonable range is to use 10-30k hours divided by the corpus duration for stage 1, and 50-75\% of that for stage 2.

\section{Experiments}
\vspace{-1mm}
\subsection{Setups}
Detailed experiments are conducted on the 960h LBS corpus \cite{libsp} and the 300h SWB corpus \cite{swb}. 
We evaluate the proposed training pipeline on context-1 transducer models using phonemes for both corpora, and full-context transducer models using 5k acoustic data-driven subword modeling (ADSM) units \cite{zhou2021ADSM} for LBS. 
Additionally, we reduce the LBS phoneme inventory in the official lexicon by unifying stressed phonemes, e.g. (AA0, AA1, AA2): AA, which we find harmless for the overall performance.
Following \cite{zhou2021phonemeTransducer}, we further apply end-of-word augmentation to the phoneme sets of both corpora.
For SWB, we use Hub5'00 as dev set, and Hub5'01 and RT'03 as test sets.

Similar as \cite{zeineldeen2022hybridconformer}, we use a $12 \times 512$ conformer (Conf.) \cite{Gulati20conformer} encoder with an initial VGG network \cite{simonyan2015vgg}.
The VGG network contains 3 3-by-3 convolutional layers with number of filters 32, 64 and 64, respectively. The last 2 convolutional layers use a temporal stride of 2 to achieve a total subsampling of factor 4. Additionally, we swap the convolution and multi-headed self-attention modules in the conformer blocks as this gives slightly better performance in our setup. For the prediction network, we use $2 \times 640$ feed-forward network and $2 \times 640$ LSTM layers for context-1 and full-context transducer models, respectively. The standard additive joint network is used, which contains a linear-tanh layer of size 1024 and a final linear-softmax layer.

We use gammatone features (LBS: 50-dim; SWB: 40-dim) \cite{schluter2007gt}, and specaugment \cite{zoph2019specaugment} except for stage 3. 
All model training is performed on single GTX 1080 Ti GPU.
By default, we apply 1-pass LM SF decoding, where word-level LM is used for phoneme transducers.
The word-level transformer (Trafo) LMs are the same as in \cite{irie19trafolm} for LBS and \cite{irie2019asru} (sentence-wise) for SWB, while the ADSM Trafo LM is the same as in \cite{zhou2022transducerILM}.

\begin{table}[t!]
\caption{\it Number of epochs (\#ep) for different transducer models in stage 1 and 2 training, and corresponding WER [\%] results on the LBS dev-clean/other sets and SWB Hub5'00 set. All results with LM are from shallow fusion (SF) decoding only.}
\vspace{-1.5mm}
\scalebox{0.9}{\parbox{1\linewidth}{%
\setlength{\tabcolsep}{0.23em}
\begin{center}\label{tab:stage1-2}
\begin{tabular}{|c|c|c|c|c||c|c||c|c|c|c|}
\hline
\multirow{3}{*}{\shortstack[c]{Train\\Stage}} & \multicolumn{6}{c||}{Phoneme Transducer} & \multicolumn{4}{c|}{ADSM Transducer} \\ \cline{2-11}
 & \multirow{2}{*}{LM} & \multicolumn{3}{c||}{LBS} & \multicolumn{2}{c||}{SWB} & \multirow{2}{*}{LM} & \multicolumn{3}{c|}{LBS} \\ \cline{3-7} \cline{9-11}
 & & \#ep & clean & other & \#ep & {\scriptsize Hub5'00} & & \#ep & clean & other \\ \hline
1 & \multirow{2}{*}{4gram} & 20 & 2.9 & 6.9 & 50 & 11.4 & \multirow{2}{*}{-} & 30 & 3.1 & 8.4 \\ \cline{1-1} \cline{3-7} \cline{9-11}
\multirow{2}{*}{2} & & \multirow{2}{*}{15} & 2.6 & 6.0 & \multirow{2}{*}{36} & 10.7 &  & \multirow{2}{*}{15} & 2.7 & 7.3 \\ \cline{2-2} \cline{4-5} \cline{7-8} \cline{10-11}
 & Trafo & & 1.8 & 4.1 & & 9.9 & Trafo & & 1.9 & 4.6 \\ \hline
\end{tabular}
\end{center}}}
\vspace{-0.5mm}
\end{table}

\begin{table}[t!]
\caption{\it Stage-1 Viterbi training using various CTC models for alignment generation and  correspondingly-trained phoneme transducer models (all 20 epochs). WER [\%] results using 4gram LM SF decoding on the LBS dev-other set.}
\vspace{-1.5mm}
\scalebox{0.85}{\parbox{1\linewidth}{%
\setlength{\tabcolsep}{0.43em}
\begin{center}\label{tab:stage1}
\begin{tabular}{|c|c|c|c||c|c|c|}
\hline
\multicolumn{4}{|c||}{Alignment CTC Model} & \multicolumn{3}{c|}{ Stage-1 Phoneme Transducer } \\ \hline
NN & size & \#ep & WER & Encoder NN & size & WER \\ \hline
\multirow{4}{*}{BLSTM} & \multirow{4}{*}{$6\times512$} & \multirow{3}{*}{20} & \multirow{3}{*}{9.7} & \multirow{2}{*}{BLSTM} & $6\times512$ & 8.1 \\ \cline{6-7}
 & & & & & $6\times640$ &  7.7 \\ \cline{5-7}
 & & & & \multirow{5}{*}{VGG-Conf.} & \multirow{5}{*}{12 blocks} & 6.9 \\ \cline{3-4} \cline{7-7}
 & & \multirow{4}{*}{30} & 9.4 & & & 6.9 \\ \cline{1-2} \cline{4-4} \cline{7-7}
\multirow{2}{*}{\shortstack[c]{BLSTM-\\Conf.}} & \multirow{2}{*}{\shortstack[c]{$3\times512$\\6 blocks}} &  & \multirow{2}{*}{9.0} & & & \multirow{2}{*}{6.9} \\ 
 & & & & & &  \\ \cline{1-2} \cline{4-4} \cline{7-7}
VGG-Conf. & 12 blocks &  & 7.6 & & & 6.9 \\ \hline
\end{tabular}
\end{center}}}
\vspace{-4mm}
\end{table}

\subsection{Stage 1: Viterbi training}
By default, we use a $6 \times 512$ BLSTM for the alignment CTC model, which adopts the same subsampling via max-pooling layers in the middle.
For LBS and SWB, we train the CTC model for 20 epochs and 35 epochs, respectively.
After generating the Viterbi alignment, we apply the stage-1 training on transducer models.
The $\text{LR}_\text{peak}$ for phoneme and ADSM transducers are $8e^{-4}$ and $3e^{-4}$, respectively.
The corresponding epochs and word error rate (WER) results are shown in \Cref{tab:stage1-2}.

We use the LBS phoneme transducer to illustrate the robustness of this training stage. 
Different CTC models are evaluated for alignment generation, which covers different NN structures, epochs and accuracy as shown in \Cref{tab:stage1}. 
We see that the correspondingly Viterbi-trained conformer-transducer models all achieve the same performance, where the same recipe also works well for BLSTM-transducer models.
We believe the cost of the alignment CTC model can be further reduced, which we didn't investigate much here.

\begin{table}[t!]
\caption{\it Efficiency illustration (on single GTX 1080 Ti GPU) of the proposed training with phoneme transducer; vs. standard training under similar WER [\%] on the LBS dev sets}
\vspace{-1.5mm}
\scalebox{0.85}{\parbox{1\linewidth}{%
\setlength{\tabcolsep}{0.22em}
\begin{center}\label{tab:stage2}
\begin{tabular}{|l|l|c|c|c|c||l|c|c|}
\hline
\multirow{2}{*}{Model Train} & \multicolumn{5}{c||}{LBS} & \multicolumn{3}{c|}{SWB} \\ \cline{2-9}
& \#ep & hour/ep & $\sum$ hour & clean & other & \#ep & hour/ep & $\sum$ hour \\ \hline
standard    & 40 & 18.3 & 732 & 2.6 & 6.1 & \multicolumn{3}{c|}{-} \\ \hline
stage 1     & 20 & 6    &  \multicolumn{3}{c||}{-} & 50 & 1.8 & -\\ \cline{4-6} \cline{9-9}
+ stage 2   & + 15 & 18.3 & 394.5 & 2.6 & 6.0 & + 36 & 5.5 & 288 \\ \hline
\end{tabular}
\end{center}}}
\vspace{-0.5mm}
\end{table}

\begin{table}[t!]
\caption{\it WER [\%] and LM integration effect of stage 3 training on the stage-2 transducer models; results evaluated with SF and ILM correction on the LBS dev-other and SWB Hub5'00 sets; further detailed scales and Sub/Del/Ins error rate [\%] on LBS.}
\vspace{-1.5mm}
\scalebox{0.9}{\parbox{1\linewidth}{%
\setlength{\tabcolsep}{0.35em}
\begin{center}\label{tab:stage3}
\begin{tabular}{|c|c|c|c|c|c|c|c|c||c|}
\hline
\multirow{3}{*}{Model} & \multirow{3}{*}{\shortstack[c]{Train\\Stage}} & \multirow{3}{*}{LM} & \multicolumn{6}{c||}{LBS} & SWB \\ \cline{4-10}
 & & & \multirow{2}{*}{${\lambda}_1$} & \multirow{2}{*}{${\lambda}_2$}  & \multicolumn{4}{c||}{dev-other} & {\footnotesize Hub5'00} \\ \cline{6-10}
 & & &  & & WER & {Sub} & {Del} & {Ins} & WER \\ \hline
\multirow{4}{*}{\shortstack[c]{phoneme\\Transd.}} & \multirow{2}{*}{2} & \hspace{-2mm} Trafo & 0.9 & 0 & 4.1 & 3.1 & 0.6 & 0.4 &  9.9 \\ 
& & + ILM & 1.0 & 0.2 & 3.9 & 3.1 & 0.4 & 0.4 & 9.7 \\ \cline{2-10}
& \multirow{2}{*}{3} & \hspace{-2mm} Trafo & 1.3 & 0 & \bf3.7 & 3.0 & 0.4 & 0.4 & 9.3 \\ 
& & + ILM & 1.4 & 0.1 & \bf3.7 & 2.9 & 0.4 & 0.4 & \bf9.2 \\ \hline
\multirow{4}{*}{\shortstack[c]{ADSM\\Transd.}} & \multirow{2}{*}{2} & \hspace{-2mm} Trafo & 0.6 & 0 & 4.6 & 3.5 & 0.7 & 0.4 & \multirow{4}{*}{-} \\ 
& & + ILM & 0.8 & 0.4 & 4.0 & 3.2 & 0.4 & 0.4 & \\ \cline{2-9}
& \multirow{2}{*}{3} & \hspace{-2mm} Trafo & 0.9 & 0 & 4.2 & 3.4 & 0.4 & 0.4 & \\ 
& & + ILM & 1.2 & 0.3 & 4.0 & 3.2 & 0.4 & 0.4 & \\ \hline
\end{tabular}
\end{center}}}
\vspace{-4.5mm}
\end{table}

\subsection{Stage 2: full-sum fine-tuning}
\vspace{-0.5mm}
Based on cross-validation score or intermediate recognition, the best model checkpoints of stage 1 are selected for stage-2 training.
For SWB, we increase dropout to 0.2 from this stage on to avoid overfitting with the fine-tuning process.
We find $\text{LR}_\text{peak}$ around $5e^{-5}$ to be a good optimum for all models.
The corresponding number of epochs and WER results of this training stage are shown in \Cref{tab:stage1-2}. We see that decent improvements are achieved in all cases with only a small amount of epochs.

In \Cref{tab:stage2}, we show the efficiency of the proposed training pipeline using phoneme transducers, which is compared with the standard pipeline on LBS. 
We use the well-trained VGG-Conf. CTC model from \Cref{tab:stage1} for encoder initialization and directly train a transducer model using \Cref{eq:fullsum}.  
The same OCLR schedule as in stage 1 is applied.
We train this standard model for 40 epochs till it reaches the same performance as our stage-2 model. Based on the total time used, our proposed pipeline achieves a 46\% relative speed-up.

\subsection{Stage 3: fast sequence training for LM integration}
\vspace{-0.5mm}
The best models from stage 2 are then further trained in stage 3 with the fast MBR training described in \Cref{sec:fastMBR}.
As the weaker LM for lattice generation, we use the same recognition 4gram LM for phoneme transducers, and a $1\times1024$ LSTM LM trained on the LM text data for the ADSM transducer.
We filter out utterances longer than 16s in time, or containing more than 190 phonemes or 90 ADSM labels in the transcription.
The MBR training uses an $N$-best of size 4.
Additionally for the SWB phoneme transducer, we exclude the non-speech labels for sequence uniqueness and risk computation. 
Besides WER, we also evaluate LM integration effect using SF and ILM correction for decoding, where the zero-encoder ILM approach \cite{variani2020hat, ILME} is applied for the latter. The detailed results are shown in \Cref{tab:stage3}.

With LM SF decoding, consistent improvements are achieved by the stage-3 training upon all the stage-2 transducer models. 
As claimed in \Cref{sec:fastMBR}, we see that the higher Del errors are reduced to a balanced level with Ins, while the LM scale ${\lambda}_1$ is increased to a more reasonable value to further suppress the Sub errors.
With ILM correction, the stage-2 transducer models already achieve better results.
Yet the stage-3 training brings further 5\% relative improvements to both phoneme transducers.
However, the benefit of ILM correction for the full-context ADSM transducer becomes much smaller after stage-3 training, which leads to no overall improvement comparing to the stage-2 ADSM transducer with ILM correction.
It seems that the proposed fast sequence training is more helpful when the ILM is less powerful, e.g. limited context. We will further investigate this joint effect in future work.


\begin{table}[t!]
\caption{\it Overall WER [\%] results on LBS vs. literature including number of parameters (\#pm) and epochs (\#ep)}
\vspace{-1.5mm}
\scalebox{0.9}{\parbox{1\linewidth}{%
\setlength{\tabcolsep}{0.25em}
\begin{center}\label{tab:libsp}
\begin{tabular}{|c|c|c|c|c|c|c|c|c|}
\hline
\multirow{2}{*}{Work} & \multicolumn{3}{c|}{Model} & \multirow{2}{*}{LM} & \multicolumn{2}{c|}{dev} & \multicolumn{2}{c|}{test} \\ \cline{2-4} \cline{6-9}
 & Approach & \#pm & \#ep & & clean & other & clean & other \\ \hline
\cite{park2020specaugAdapt} & RNN AED & 360M & 600 & LSTM & \multicolumn{2}{c|}{\multirow{4}{*}{-}} & 2.2 & 5.2 \\ \cline{1-5} \cline{8-9}
\cite{wang2020trafoHybrid} & Trafo Hybrid & 81M & 100 & \multirow{2}{*}{Trafo} & \multicolumn{2}{c|}{} & 2.3 & 4.9 \\ \cline{1-4} \cline{8-9}
\cite{zhang2020wpmHybrid} & Trafo CTC & 124M & 200 & & \multicolumn{2}{c|}{} & 2.1 & 4.2 \\ \cline{1-5} \cline{8-9}
\cite{Gulati20conformer} & Conf. Transd. & 119M & - & LSTM & \multicolumn{2}{c|}{} & 1.9 & 3.9 \\ \hline \hline
\multirow{3}{*}{this} & ADSM Transd. & 87M & 45 & Trafo & 1.8 & 4.0 & 1.9 & 4.4 \\ \cline{2-9}
& \multirow{2}{*}{\shortstack[c]{phoneme\\Transd.}} & \multirow{2}{*}{75M} & \multirow{2}{*}{35} & 4gram & 2.5 & 5.7 & 2.9 & 6.2 \\ \cline{5-9}
& & & & Trafo & 1.7 & 3.7 & 2.1 & 4.1 \\ \hline
\end{tabular}
\end{center}}}
\vspace{-0.5mm}
\end{table}

\begin{table}[t!]
\caption{\it Overall WER [\%] results on SWB vs. literature.\\ 
{\footnotesize (*: cross-utterance evaluation with combined LSTM and Trafo LMs)}}
\vspace{-1.5mm}
\scalebox{0.9}{\parbox{1\linewidth}{%
\setlength{\tabcolsep}{0.3em}
\begin{center}\label{tab:swb}
\begin{tabular}{|c|c|c|c|c|c|c|c|}
\hline
\multirow{2}{*}{Work} & \multicolumn{3}{c|}{Model} & \multirow{2}{*}{LM} & \multirow{2}{*}{\footnotesize Hub5'00} & \multirow{2}{*}{\footnotesize Hub5'01} & \multirow{2}{*}{\footnotesize RT'03} \\ \cline{2-4}
 & Approach & \#pm & \#ep &  &  & &   \\ \hline
\cite{zeineldeen2022hybridconformer} & Conf. Hybrid & 88M & - & Trafo & 10.3 & 9.7 & - \\ \hline
\cite{saon21rnnt} & RNN Transd. & 57M & 100 & \multirow{2}{*}{LSTM} & 9.7 & 10.1 & 12.6 \\ \cline{1-4} \cline{6-8}
\cite{zoltan2020swb} & RNN AED & 280M & 250 & & 9.8 & 10.1 & 12.0 \\ \hline
\cite{tuske2021swbAED} & Conf. AED & 68M & 250 & + Trafo & \hspace{1.5mm}$8.4^{*}$ & \hspace{1.5mm}$8.5^{*}$ & \hspace{1.5mm}$9.9^{*}$ \\ \hline \hline
\multirow{2}{*}{this} & \multirow{2}{*}{\shortstack[c]{phoneme\\Transd.}} & \multirow{2}{*}{75M} & \multirow{2}{*}{86} & 4gram & 10.3 & 10.6 & 11.8 \\ \cline{5-8}
& & & & Trafo & 9.2 & 9.4 & 10.5 \\ \hline
\end{tabular}
\end{center}}}
\vspace{-4.2mm}
\end{table}

\subsection{Overall performance}
\vspace{-0.5mm}
Finally, we evaluate the overall performance of our best phoneme transducer models from stage 3 and best ADSM transducer model from stage 2.
We switch the ILM correction for ADSM transducer to the mini-LSTM ILM approach \cite{zeineldeen2021ilm} as done in \cite{zhou2022transducerILM}.
All scales are optimized on the dev sets.
We compare these results with other top systems from the literature in \Cref{tab:libsp} and \Cref{tab:swb} for LBS and SWB, respectively.
Our best systems are competitive with SOTA results, while our models are smaller and/or trained with much less epochs.

\vspace{-2mm}
\section{Conclusions}
\vspace{-0.7mm}
In this work, we presented an efficient 3-stage progressive training pipeline for both phoneme and subword neural transducer models. 
The provided detailed recipes and design principles of each stage are experimentally verified on both LBS and SWB corpora. 
Compared with the standard pipeline, our proposed training pipeline achieves a large reduction on time and computation costs as well as complexity.
This allows us to build conformer transducer models approaching SOTA performance from scratch with a single GPU in just 2-3 weeks.

\vspace{-0.6mm}
\begin{center}
\bf Acknowledgements
\end{center}
\vspace{-0.5mm}
\scriptsize
This work was partially supported by the Google Faculty Research Award for ``Label Context Modeling in Automatic Speech Recognition'', and by NeuroSys which, as part of the initiative ``Clusters4Future'', is funded by the Federal Ministry of Education and Research BMBF (03ZU1106DA). \\
We thank Zuoyun Zheng for training the ADSM LMs.

\newpage
\let\normalsize\small\normalsize
\let\OLDthebibliography\thebibliography
\renewcommand\thebibliography[1]{
        \OLDthebibliography{#1}
        \setlength{\parskip}{-0.3pt}
        \setlength{\itemsep}{1pt plus 0.07ex}
}

\bibliographystyle{IEEEtran}
\bibliography{refs}

\end{document}